\pdfoutput=1

\documentclass[11pt]{article}

\usepackage[]{EMNLP2022}

\usepackage{times}
\usepackage{latexsym}

\usepackage{graphics}
\usepackage{graphicx}
\usepackage{xurl}
\usepackage{array}
\usepackage{booktabs}
\usepackage{xcolor}

\usepackage[T1]{fontenc}

\usepackage[utf8]{inputenc}

\usepackage{microtype}

\usepackage{inconsolata}

\renewcommand\fbox{\fcolorbox{lightgray}{white}}

\title{Discovering Differences in the Representation of\\People using Contextualized Semantic Axes}

\author{
 Li Lucy, Divya Tadimeti,\and David Bamman\\
 University of California, Berkeley \\
  {\sf \{lucy3\_li, dtadimeti, dbamman\}@berkeley.edu} \\
}

\begin{document}
\maketitle
\begin{abstract}
A common paradigm for identifying semantic differences across social and temporal contexts is the use of static word embeddings and their distances. In particular, past work has compared embeddings against ``semantic axes'' that represent two opposing concepts. We extend this paradigm to BERT embeddings, and construct contextualized axes that mitigate the pitfall where antonyms have neighboring representations. We validate and demonstrate these axes on two people-centric datasets: occupations from Wikipedia, and multi-platform discussions in extremist, men's communities over fourteen years. In both studies, contextualized semantic axes can characterize differences among instances of the same word type. In the latter study, we show that references to women and the contexts around them have become more detestable over time.
\end{abstract}

\section{Introduction}

\textit{\textbf{Warning}: This paper contains content that may be offensive or upsetting.}

Quantifying and describing the nature of language differences is key to measuring the impact of social and cultural factors on text. Past work has compared English embeddings for people to adjectives or concepts \cite{garg2018word,julia_dehumanization2020,charlesworth2022}, or projected embeddings against axes representing contrasting attributes \cite{turney2003measuring,an-etal-2018-semaxis,kozlowski2019geometry,field-tsvetkov-2019-entity,mathew2020polar,kwak2021frameaxis,lucy-bamman-2021-gender,fraser-etal-2021-understanding,grand2022semantic}. Static representations for the same word can also be juxtaposed across corpora that reflect different time periods \cite{gonen-etal-2020-simple,hamilton-etal-2016-cultural}. This paradigm of using embedding distances to uncover socially meaningful patterns has also transferred over to studies that measure biases in contextualized embeddings, such as \citet{wolfe-caliskan-2021-low}'s finding that BERT embeddings of less frequent minority names are closer to words related to unpleasantness. 

The use of ``semantic axes'' is enticing in that it offers an interpretable measurement of word differences beyond a single similarity value \cite{turney2003measuring,an-etal-2018-semaxis,kozlowski2019geometry,kwak2021frameaxis}. Words are projected onto axes where the poles represent antonymous concepts (such as \textit{beautiful}--\textit{ugly}), and the projected embedding's location along the axis indicates how similar it is to either concept. Semantic axes constructed using static, type-based embeddings have been used to analyze socially meaningful differences, such as words' associations with class \cite{kozlowski2019geometry}, or gender stereotypes in narratives \cite{huang-etal-2021-uncovering-implicit,lucy-bamman-2021-gender}.

\begin{figure}[t]
    \includegraphics[width=\columnwidth]{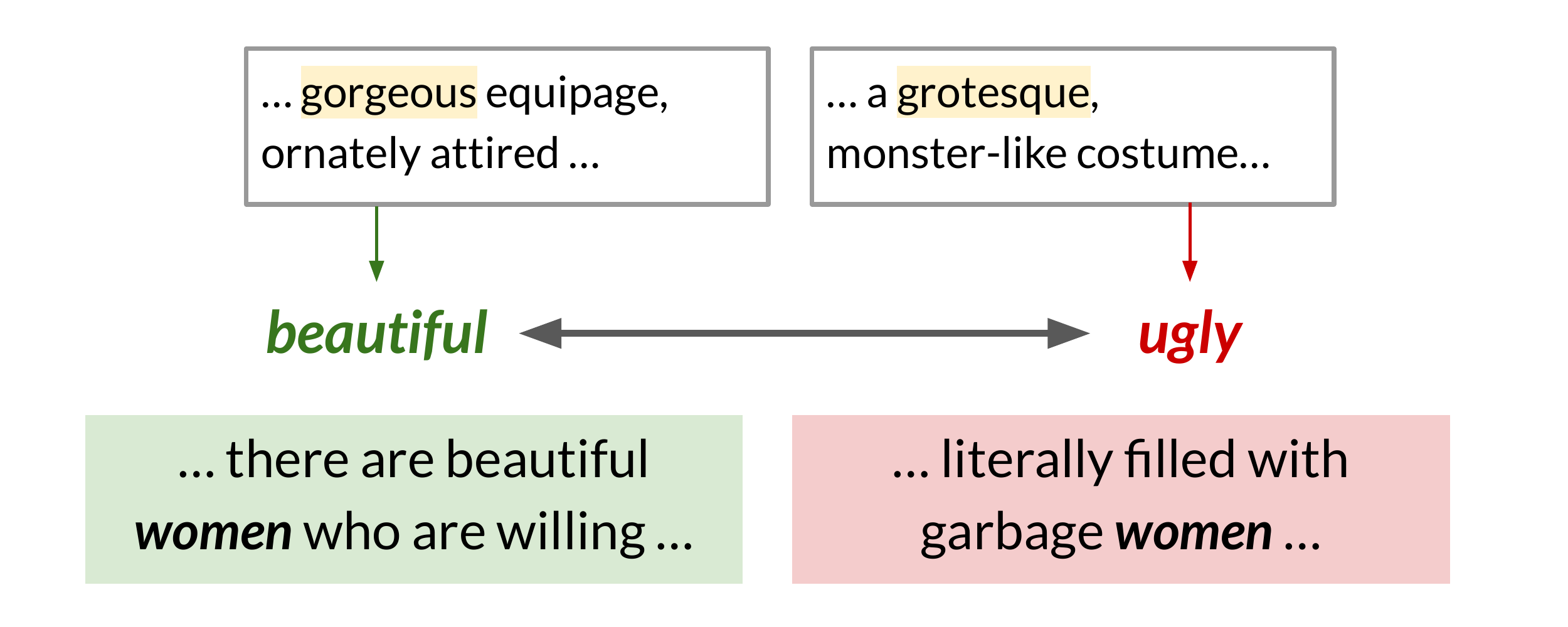}
    \centering
    \caption{An axis is constructed using embeddings of adjectives in selected contexts. These contexts are predictive of synonyms, but not antonyms, of the target adjective during masked language modeling. Token-level embeddings for people are then projected onto this axis.}
	\label{fig:intro}
\end{figure}

Our work investigates the extension and application of semantic axes to contextualized embeddings. We present a novel approach for constructing semantic axes with English BERT embeddings (Figure~\ref{fig:intro}). These axes are built to encourage self-consistency, where antonymous poles are less conflated with each other. They are able to capture semantic differences across word types as well as variation in a single word across contexts. Their ability to differentiate contexts makes them suitable for studying how a word changes across domains or across individual sentences. These axes are also more self-consistent and coherent than ones created using GloVe and other baseline approaches. 

We demonstrate the use of contextualized axes on two datasets: occupations from Wikipedia, and people discussed in misogynistic online communities. We use the former as a case where terms appear in definitional contexts, and characteristics of people are well-known. In the latter longitudinal, cross-platform case study, we examine lexical choices made by communities whose attitudes towards women tend to be salient and extreme. We chose this set of online communities as a substantive use case of our method, in light of recent attention in web science on analyzing online extremism and hate at scale \citep[e.g.][]{ribeiro2021platform,ribeiro2021evolution,aliapoulios_2021_large}. There, we analyze language change and variation along axes through a sociolinguistic lens, emphasizing that speakers use language that reflects their social identities and beliefs \cite{ch-wang-jurgens-2021-using,huffaker2017gender,card-etal-2016-analyzing,lakoff2006framing}. 

Our code, vocabularies, and other resources can be found in our Github repo:
\url{https://github.com/lucy3/context_semantic_axes}.

\section{Constructing semantic axes}\label{sec:construct}

\textbf{Static embeddings.} Several formulae for calculating the similarity of a target word to two sets of pole words have been proposed in prior work on static semantic axes. These differ in whether they take the difference between a target word's similarities to each pole \cite{turney2003measuring}, calculate a target word's similarity to the difference between pole averages \cite{an-etal-2018-semaxis,kwak2021frameaxis}, or calculate a target word's similarity to the average of several word pair differences that represent the same antonymous relationship \cite{kozlowski2019geometry}. We build on the approach of \citet{an-etal-2018-semaxis} and \citet{kwak2021frameaxis}, because it does not require us to curate multiple paired antonyms for each axis, and it draws out the difference between two concepts before a target word is compared to them, rather than after. We define an axis $V$ containing antonymous sets of adjective vectors, $S_l = \{l_1, l_2, l_3, ..., l_n\}$ and $S_r= \{r_1, r_2, r_3, ..., r_m\}$, as the following: 
$$V = \frac{1}{n}\sum_{i=1}^n l_i - \frac{1}{m}\sum_{j=1}^m r_j.$$

Relying on single-word poles for axes can be unstable to the choice of each word \cite{an-etal-2018-semaxis,antoniak-mimno-2021-bad}. \citet{an-etal-2018-semaxis} creates a pole's set of words using the nearest neighbors of a seed word, which may risk conflating unintended meanings or antonymous neighbors \cite{mrksic-etal-2016-counter,sedoc-etal-2017-semantic}. For example, one axis uses the opposite seed words \textit{green} and \textit{experienced}, but \textit{green}'s nearest neighbors include \textit{red} rather than \textit{inexperienced}. Instead of using this nearest neighbors approach, we construct poles using WordNet antonym relations. Each end of an axis aggregates synonymous and similar lemmas in WordNet synsets, which are expanded using the \textit{similar to} relation \cite{miller-1992-wordnet}. 

Our type-based embedding baseline, \textsc{glove}, uses 300-dimensional GloVe vectors pretrained on Wikipedia and Gigaword \cite{pennington-etal-2014-glove}. We only keep poles where both sides have at least three adjectives that appear in the GloVe vocabulary, and we also exclude acronyms, which are often more ambiguous in meaning. We start with 723 axes, where poles have on average 9.63 adjectives each. 

\textbf{Contextualized embeddings.} Static embeddings, however, present a number of limitations. Such embeddings cannot easily handle polysemy or homonymy \cite{wiedemann-etal-2019-does}, and even when they are trained on different social or temporal contexts, they require additional steps to be aligned \cite{gonen-etal-2020-simple}. Context-specific embeddings also need enough training examples of target words to create usable representations. These limitations prevent the analysis of token-based semantic variation, such as measuring how one mention of a word is more or less \textit{beautiful} than another. Our main contribution of contextualized axes uses the same WordNet-based formulation as our GloVe baseline. Rather than each word in $S_l$ or $S_r$ being represented by a single GloVe embedding, we obtain BERT embeddings over multiple occurrences of each adjective. We use BERT-base, as this model is small enough for efficient application on large datasets and is popular in previous work on semantic change and differences \citep[e.g.][]{hu-etal-2019-diachronic, lucy-bamman-2021-characterizing, giulianelli-etal-2020-analysing, zhou-etal-2022-richer, coll-ardanuy-etal-2020-living,martinc-etal-2020-leveraging}. It is also used in tutorials for researchers outside of NLP, which means it has high potential use in computational social science and cultural analytics \cite{berthumanists2022}. 

For contextualized axes, we obtain a potential pool of contexts for adjectives sampled over all of Wikipedia from December 21, 2021, preprocessed using \citet{Wikiextractor2015}'s text extractor. This sample contains up to 1000 sentences, or contexts, that contain each adjective, and we avoid contexts that are too short (over 10 tokens) or too long (over 150 tokens).\footnote{This length cutoff made the data more manageable, and 90\% of BERT's training steps were originally on 128-length sequences \cite{devlin-etal-2019-bert}.}

We experiment with two methods of obtaining contextualized BERT embeddings for each adjective: a random ``default"  (\textsc{bert-default}) and one where contexts are picked based on word probabilities (\textsc{bert-prob}). For \textsc{bert-default}, we take a random sample of 100 contextualized embeddings across the adjectives in each pole. Since words can be nearest neighbors with their antonyms in semantic space \cite{mrksic-etal-2016-counter,sedoc-etal-2017-semantic}, our main approach, \textsc{bert-prob}, aggregates word embeddings over contexts that highlight contrasting meanings of axes' poles. 

To select contexts, we mask out the target adjective in each of its 1000 sentences, and have BERT-base predict the probabilities of synonyms and antonyms for that masked token. We remove contexts where the average probability of antonyms is greater than that of synonyms, sort by average synonym probability, and take the top 100 contexts. One limitation of our approach is that predictions are restricted to adjectives that can be represented by one wordpiece token. If none of the words on a pole of an axis appear in BERT's vocabulary, we backoff to \textsc{bert-default} to represent that axis.

For each axis type, we also have versions where words' embeddings are $z$-scored, which has been shown to improve BERT's alignment with humans' word similarity judgements \cite{timkey-van-schijndel-2021-bark}. For $z$-scoring, we calculate mean and standard deviation BERT embeddings from a sample of around 370k whole words from Wikipedia. As recommended by \citet{bommasani-etal-2020-interpreting}, we use mean pooling over wordpieces to produce word representations when necessary, and we extend this approach to create bigram representations as well. These embeddings are a concatenation of the last four layers of BERT, as these tend to capture more context-specific information \cite{ethayarajh-2019-contextual}. 

\section{Internal validation}

We internally validate our axes for self-consistency. For each axis, we remove one adjective's embeddings from either side, and compute its cosine similarity to the axis constructed from the remaining adjectives. For BERT approaches, we average the adjective's multiple embeddings to produce only one before computing its similarity to the axis. In a ``consistent'' axis, a left-out adjective should be closer to the pole it belongs to. That is, if it belongs to $S_l$, its similarity to the axis should be positive. We average these leave-one-out similarities for each pole, negating the score when the adjective belongs to $S_r$, to produce a consistency metric, $\mathcal{C}$. Table~\ref{tbl:consist} shows $\mathcal{C}$ for different axis-building methods.\footnote{We assign $\mathcal{C}$ to 0 if only one unique adjective's contexts are chosen to create a pole for \textsc{bert-prob}, because in that case, we are unable to run the leave-one-out test for that pole.} An axis is ``consistent'' if both of its poles have $\mathcal{C} \geq 0$. 

\textsc{glove}'s most inconsistent axis poles often involve directions, such as \textit{east $\leftrightarrow$ west}, \textit{left-handed $\leftrightarrow$ right-handed}, and \textit{right $\leftrightarrow$ left}. These concepts may be difficult to learn from text without grounding. We find that the various BERT approaches' most inconsistent axes include direction-related ones as well, but they also struggle to separate concepts such as \textit{lower-class $\leftrightarrow$ upper-class}. 

The best method for producing consistent axes is $z$-scored \textsc{bert-prob}, with a significant difference in $\mathcal{C}$ from $z$-scored \textsc{bert-default} and \textsc{glove} (Mann-Whitney U-test, $p<0.001$). It also produces the highest number of consistent axes. \textsc{Glove} presents itself as a formidable baseline,\footnote{We also tried $z$-scoring \textsc{glove} embeddings, but this worsened internal consistency ($\mathcal{C} = 0.098$).} and \textsc{bert-default} struggles in comparison to it. 

\begin{table}[t]
\centering
\resizebox{0.9\columnwidth}{!}{
\begin{tabular}{l  c  c}
\toprule
\textbf{Method} & \textbf{Average $\mathcal{C}$} & \textbf{\# of consistent axes} \\
\midrule
\textsc{glove}& 0.101 (0.006)& 503\\
\textsc{bert-default} & 0.084 (0.006)& 393\\
\textsc{bert-default}$^z$ & 0.111 (0.007) & 468\\
\textsc{bert-prob}& 0.101 (0.006) & 436\\
\textbf{\textsc{bert-prob}$^z$} & \textbf{0.133 (0.007)} & \textbf{512}\\
\bottomrule
\end{tabular}}
\caption{A table of $\mathcal{C}$, averaged across poles, with 95\% confidence intervals (CI) in parentheses. The $z$ symbol represents $z$-scored approaches.}
\label{tbl:consist}
\end{table}

\begin{table*}[t]
\centering
\resizebox{\linewidth}{!}{
\begin{tabular}{l | l  l | l  l }
\toprule
{\large\textbf{Category}} & \multicolumn{2}{c}{\large\textbf{Occupation Experiment}} & \multicolumn{2}{|c}{\large\textbf{\textit{Person} Experiment}} 
\\\midrule
Writing & creative, fanciful, fictive & formal, logical, discursive & 
+ folksy, unceremonious, casual & + ignoble, common, plebeian \\
Entertainment & transcribed, taped, recorded & structural, constructive, creative & + trademarked, branded, copyrighted & + emotional, soupy, slushy\\
Art & unostentatious, aesthetic, artistic & creative, fanciful, fictive & + activist, active, hands-on & + practiced, proficient, adept \\
Health & unhealthy, pathologic, asthmatic & rehabilitative, structural, constructive & + confirmable, empirical, experiential & + teetotal, dry, drug-free\\
Agriculture & drifting, mobile, unsettled & rustic, agrarian, bucolic & + boneless, deboned, boned & - rehabilitative, structural, constructive\\
Government & amenable, answerable, responsible & policy-making, political, governmental & + respectful, deferential, honorific & + amenable, answerable, responsible \\
Sports & spry, gymnastic, sporty & zealous, ardent, enthusiastic & - amenable, answerable, responsible & - subject, subservient, dependent\\
Engineering & formal, logical, discursive & rehabilitative, structural, constructive& + coeducational, integrated, mixed & + advanced, high, graduate \\
Science & humanistic, humane, human-centered & zealous, ardent, enthusiastic & + humanistic, humane, human-centered& + stoic, unemotional, chilly\\
Math \& statistics & enumerable, estimable, calculable & formal, logical, discursive & + enumerable, estimable, calculable& - amenable, answerable, responsible\\
Social Sciences & humanistic, humane, human-centered& relational, relative, comparative & + significant, portentous, probative & + humanistic, humane, human-centered\\
\bottomrule
\end{tabular}}
\caption{The top two $z$-scored \textsc{bert-prob} axis poles, ordered from left to right, for each occupation category and experiment. Each pole is represented by three example adjectives drawn from the set used to construct that pole. Since the \textit{person} experiment compares each occupation category to all others, + or - indicates the direction of the shift in axis similarity. For example, sports occupations are still closer to \textit{responsible} than \textit{irresponsible}, just less so (-) than other occupations.}
\label{tbl:top_axes_occ}
\end{table*}

\section{External validation}

Previous work on static semantic axes validates them using sentiment lexicons, exploratory analyses, and human-reported associations \cite{an-etal-2018-semaxis, kwak2021frameaxis, kozlowski2019geometry}. We perform external validation of self-consistent axes on a dataset where people appear in a variety of well-defined and known contexts: occupations from Wikipedia. We conduct two main experiments. In the first, we test whether contextualized axes can detect differences across occupation terms, and in the second, we investigate whether they can detect differences across contexts. 

\subsection{Data}

We collect eleven categories of unigram and bigram occupations from Wikipedia lists: Writing, Entertainment, Art, Health, Agriculture, Government, Sports, Engineering, Science, Math \& Statistics, and Social sciences (Appendix~\ref{appdx:wiki}). The number of occupations per category ranges from 3 in Math \& Statistics to 48 in Entertainment, with an average of 27.2. We use the MediaWiki API to find Wikipedia pages for occupations in each list if they exist and follow redirects when necessary (e.g. \textit{Blogger} redirects to \textit{Blog}). For each occupation's singular form, we extract sentences in its page that contains it. In total, we have 3,015 sentences for 300 occupations.

\subsection{Term-level experiment (occupations)}

Each occupation is represented by a pre-trained GloVe embedding or a BERT embedding averaged over all occurrences on its page. If an axis uses $z$-scored adjective embeddings, we also $z$-score the occupation embeddings compared to it. We assign poles to occupations based on which side of the axis they are closer to via cosine similarity. Top poles are highly related to their target occupation category, as seen by the examples for $z$-scored \textsc{bert-prob} in Table~\ref{tbl:top_axes_occ}. 

One limitation for interpretability is that word embeddings' proximity can reflect any type of semantic association, not just that a person actually \textit{has} the attributes of an adjective. For example, adjectives related to \textit{unhealthy} are highly associated with Health occupations, which can be explained by doctors working in environments where unhealthiness is prominent. Therefore, embedding distances only provide a foggy window into the nature of words, and this ambiguity should be considered when interpreting word similarities and their implications. This limitation applies to both static embeddings and their contextualized counterparts. 

\begin{table}[t]
\centering
\resizebox{\columnwidth}{!}{
\begin{tabular}{l  c  c}
\toprule
\textbf{Method} & \textbf{Occupation Experiment} & \textbf{\textit{Person} Experiment} \\
\midrule
\textsc{glove}& 3.485 ($\pm$ 0.491) & - \\
\textsc{bert-default} & 3.576 ($\pm$ 0.429)& 2.697 ($\pm$ 0.361) \\
\textsc{bert-default}$^z$ & 2.636 ($\pm$ 0.459)& 2.485 ($\pm$ 0.367)\\
\textsc{bert-prob}& 3.333 ($\pm$ 0.473)& 2.667 ($\pm$ 0.363)\\
\textbf{\textsc{bert-prob}$^z$} & \textbf{1.970} ($\pm$ 0.297)& \textbf{2.152} ($\pm$ 0.404)\\
\bottomrule
\end{tabular}}
\caption{Average rank of each axis-building method for each experiment, across human evaluators and occupation categories. 95\% CI in parentheses.}
\label{tbl:human_eval}
\end{table}

We conduct human evaluation on this task of using semantic axes to differentiate and characterize occupations. Three student annotators examined the top three poles retrieved by each axis-building approach and ranked these outputs based on semantic relatedness to occupation categories (Appendix~\ref{sec:human_eval}). These annotators had fair agreement, with an average Kendall's $W$ of 0.629 across categories and experiments. Though \textsc{glove} is a competitive baseline, $z$-scored \textsc{bert-prob} is the highest-ranked approach overall (Table~\ref{tbl:human_eval}). This suggests that more self-consistent axes also produce measurements that better reflect human judgements of occupations' general meaning. 

\subsection{Context-level experiment (\textit{person})}\label{sec:occ_person}

The identity of a word, and prior associations learned from BERT's training data, have the potential to overpower its in-context use \cite{field-tsvetkov-2019-entity}. Thus, we may want to discount word associations originally learned by BERT when we examine the use of a target word in a narrower context. Prior work has shown that words with higher frequency in BERT's training data tend to encode more context-specific information in their embeddings \cite{ethayarajh-2019-contextual,zhou2021frequency,wolfe-caliskan-2021-low}. To investigate whether contextualized axes can measure context changes for people, we replace all occupation bigrams and unigrams with \textit{person}, a very common word. This also makes contexts across different words comparable to each other, a property which we will leverage later in Section~\ref{sec:mano_context}. 

Each \textit{person} embedding is averaged over one occupation's contexts. The identity of \textit{person} tends to overpower its similarity to axes across contexts, in that the top-ranked poles are similar across occupation categories. So, in contrast to the previous occupation experiment, additional steps are needed to draw out meaningful differences in how \textit{person} is used in one group of contexts from its typical use. To do this, we estimate the average cosine similarity to axes of $n$ \textit{person} embeddings in occupational contexts using 1000 bootstrapped samples, where $n$ is the number of terms in an occupation category. We take the axes with the highest statistically significant ($p$ < 0.001, one-sample $t$-test) difference in cosine similarity. 

We assume that occupations' Wikipedia pages mention them within definitional contexts, so top-ranked poles should reflect the original occupation replaced by \textit{person}. These top poles are less intuitive than those outputted by the earlier term-level experiment (Table~\ref{tbl:top_axes_occ}). Still, in some cases, such as for Government and Math \& Statistics occupations, we uncover relative differences that distinguish one category from others. We only show three adjectives in the top two poles in Table~\ref{tbl:top_axes_occ} due to space considerations, but moving further down the list for $z$-scored \textsc{bert-prob} uncovers additional meaningful poles. For example, the pole \textit{spry, gymnastic, sporty} is the third most prominent shift and highest similarity increase (+) in the \textit{person} experiment for Sports occupations. In addition, human evaluators preferred \textsc{bert-prob} over other approaches (Table~\ref{tbl:human_eval}, Appendix~\ref{sec:human_eval}).

\section{Measuring change and variation}

Now that we have contextualized semantic axes that can measure differences across words and contexts, we apply them onto a domain that can showcase salient and socially meaningful variation. NLP research on harmful language often employs methods that focus on the target group, such as measuring their association with other words \cite{zannettou2020quantitative,garg2018word,tahmasbi2021bat,field-tsvetkov-2019-entity}, or with biases in models \cite{wolfe-caliskan-2021-low,ghosh-etal-2021-detecting}. We illustrate the application of self-consistent $z$-scored \textsc{bert-prob} axes onto the \textit{manosphere}, which is a collection of communities with mostly male users who hold alternative beliefs around relationships and gender. We use the same axes we presented earlier, which were created using Wikipedia data, because Wikipedia provides more normative coverage of a variety of adjectives than topic-specific communities. This way, we examine how entities in the manosphere orient themselves against typical adjectival uses and meanings. 

The manosphere has been linked to acts of violence in the physical world \cite{hoffman2020violence}, and most members believe that men are systemically disadvantaged in society \cite{van2021digesting,marwick2018tears,Lin2017antifem,ging2019masculinities}. These communities focus on heterosexual relationships and masculinity, and feature a dynamic linguistic landscape. Much prior work on the manosphere has been qualitative, such as ethnographies \cite{Lin2017antifem,Lumsden2019,van2021digesting}. There have been a few quantitative analyses of their language, usually focusing on phrase and word frequencies in a few communities \cite{farrell2019exploring, gothard2021a, LaViolette_Hogan_2019,jaki2019incels}. As an example involving word vectors, \citet{farrell2020jargon} uses static embeddings identify the meanings of incels' neologisms by inspecting words' nearest neighbors. 

Our case study extends beyond prior work with its methodology and scale. We use contextualized semantic axes to tackle one question: how have references to women and contexts around them changed over fourteen years? 

\subsection{Data}\label{sec:mano_data}

We use a taxonomy of subreddits and external forums described by \citet{ribeiro2021evolution}, who show that the manosphere began with ideologies such as pick-up artists (PUA) and Men's Rights Activists (MRA), and evolved into more extreme ones such as The Red Pill (TRP), incels (short for \textit{involuntary celibate}) and Men Who Go Their Own Way (MGTOW), with users moving from older to newer ideologies. We call this dataset \textsc{extreme\_rel}, because it contains extreme views of relationships. 

We use Reddit posts and comments from March 2008 to December 2019 from subreddits listed in \citet{ribeiro2021evolution}'s study, downloaded from Pushshift \cite{baumgartner2020pushshift}. We slightly modify their taxonomy by separating out incel subreddits where the intended userbase are women (\textit{femcels}), and also include a newer set of subreddits focused on ``Female Dating Strategy" (FDS), a women-led community analogous to TRP \cite{melmag,jezebel}. Therefore, we have 60 subreddits in seven ideological categories: Incels, MGTOW, PUA, MRA, TRP, FDS, and Femcels\footnote{The women-led communities, FDS and Femcels, make up only 1.8\% of posts and comments in \textsc{extreme\_rel}'s Reddit subset.}  (Appendix~\ref{sec:appdx_communities}). This Reddit subset of \textsc{extreme\_rel} contains over 1.3 billion tokens.

We also include seven external forums provided by \citet{ribeiro2021evolution}. These public forums include A Voice for Men (AVFM), Master Pick-up Artist (MPUA) Forum, The Attraction, incels.co, MGTOW Forum, RooshV, and Red Pill Talk.\footnote{All forums are collected by \citet{ribeiro2021evolution}, available at \href{https://zenodo.org/record/4007913\#.YiqKexBKhQI}{https://zenodo.org/record/4007913\#.YiqKexBKhQI}} This forum subset of \textsc{extreme\_rel} contains over 800 million tokens spanning November 2005 to June 2019, and we remove duplicates and quoted text from posts. 

Some experiments use a subset of Reddit that shares a similar topical focus as \textsc{extreme\_rel}, but may have more mainstream views of women and relationships. We use a list\footnote{From Reddit's \href{https://www.reddit.com/r/ListOfSubreddits/wiki/listofsubreddits/\#wiki_relationships}{List of Subreddits wiki}.} of common ``Relationship'' subreddits: r/relationships, r/dating, r/relationship\_advice, r/dating\_advice, and r/breakups. We call this dataset \textsc{general\_rel}, and it contains 1.2 billion tokens from September 2009 to December 2019. For Reddit data, we do not use posts and comments written by usernames who have bot-like behavior, which we define as repeating any 10-gram more than 100 times. 

\subsection{Vocabulary}

We use a mix of NER, online glossaries, and manual inspection to curate a unique vocabulary of people (details in Appendix~\ref{sec:appdx_vocab}). This vocabulary has 2,434 unigrams and 4,179 bigrams, tokenized using BERT's tokenizer without splitting words into wordpieces \cite{devlin-etal-2019-bert,wolf-etal-2020-transformers}. These terms appear at least 500 times in \textsc{extreme\_rel}. 

\begin{figure*}[t]
\centering
\includegraphics[width=0.94\linewidth]{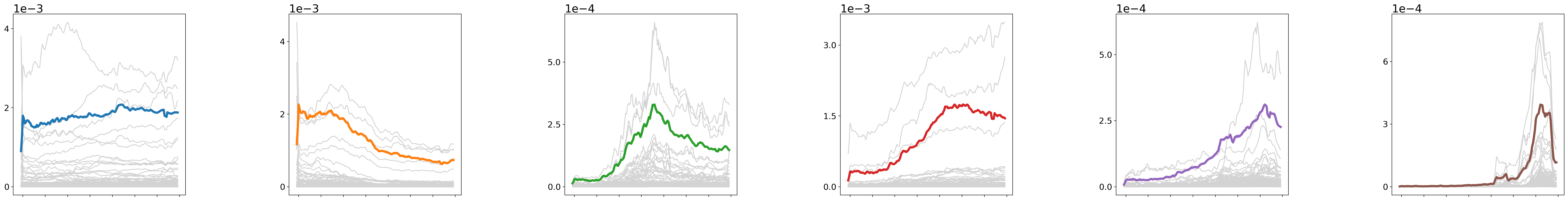}
\\
\resizebox{\linewidth}{!}{
\begin{tabular}{|p{0.17\textwidth} l|p{0.17\textwidth} l|p{0.17\textwidth} l|p{0.17\textwidth} l|p{0.17\textwidth} l|p{0.17\textwidth} l|}
\multicolumn{2}{c}{\textbf{Cluster A}} & \multicolumn{2}{c}{\textbf{Cluster B} (PUA)} & \multicolumn{2}{c}{\textbf{Cluster C} (TRP, MRA)} & \multicolumn{2}{c}{\textbf{Cluster D}} &\multicolumn{2}{c}{\textbf{Cluster E} (MGTOW)} & \multicolumn{2}{c}{\textbf{Cluster F} (Incels)} \tabularnewline
she & 1.00 & girl & 1.00 & feminist & 0.93 & women & 1.00 & females & 1.00 & foids & 0.87 \tabularnewline
female & 1.00 & girls & 1.00 & feminists & 0.93 & woman & 1.00 & virgin & 0.97 & foid & 0.87  \tabularnewline
bitch & 0.88 & chick & 1.00 & american women & 1.00 & wife & 1.00 & whore & 0.91 & thot & 0.89 \tabularnewline
girlfriend & 1.00 & chicks & 1.00 & accuser & 0.80 & mother & 1.00 & cunt & 0.77 & thots & 0.89 \tabularnewline
mom & 0.97 & gf & 1.00 & accusers & 0.80 & bitches & 0.88 & most women & 1.00 & femoids & 0.89 \tabularnewline
\end{tabular}}
\caption{Top five most frequent feminine (gender leaning > 0.75) vocabulary terms in each time series cluster, with their gender-leaning score. In each cluster's figure, cluster centers $\mu$ are thick lines, with time series of all vocab terms in light gray. Cluster centers are scaled down to half the maximum height of words' timelines. All time series start on Nov 2005 and end on Dec 2019.}
\label{tbl:vocab}
\end{figure*}

Since gender is central to the manosphere, we infer these labels based on terms' social gender in a dataset. For example, \textit{accuser} is not semantically gendered like \textit{girl} and \textit{woman}, but its social gender, estimated using pronouns, is more feminine in \textsc{extreme\_rel} than \textsc{general\_rel}. We use two stages of gender inference to account for pronoun sparsity and noise. First, we use a list of semantically gendered nouns, and second, we use feminine and masculine pronouns linked to terms via coreference resolution (details in Appendix~\ref{sec:gender_appdx}). We label each vocabulary term based on its fraction of co-occurring feminine pronouns in \textsc{extreme\_rel} and \textsc{general\_rel}, separately. We are able to label 72.5\% of the vocabulary in \textsc{extreme\_rel} and 67.0\% of it in \textsc{general\_rel}. 

\subsection{Term-level change}\label{sec:type}

\textbf{Contextualized semantic axes can reveal how word and phrase types change over time.} Here, our analyses focus on 1,482 feminine (gender-leaning > 0.75) terms in \textsc{extreme\_rel}. To capture broad snapshots of words' use, we randomly sample up to 500 sentence-level occurrences of each term in each platform and ideology (e.g. a specific forum or Reddit category) in each year. Overall $z$-scored BERT embeddings for each vocab word are averages over this stratified sample of its contexts.

The history of the manosphere is characterized by waves of different ideological communities \cite{ribeiro2021evolution}. To reflect this characterization through language, we segment our vocabulary based on when terms peak in popularity. We cluster normalized frequency time series\footnote{We smooth the time series using a moving average with a kernel size of 3, and count each term once per comment to reduce the effect of unusually long comments.} for each term using $K$-Spectral Centroid clustering (KSC) \cite{yang_leskovec_ksc}. We use their default parameters, including $K=6$. In contrast to their original approach, our symmetric distance measure $\hat{d}$ is invariant to scaling by $\alpha$ but not to the translation of the time series, so that peaks earlier in time are not clustered with those later in time: 
$$\hat{d}(x, y) = \frac{||x - \alpha y||}{||x||},$$ where $\alpha = x^Ty / ||y||^2.$

``Waves" of term types for people correspond to ideological change. Figure~\ref{tbl:vocab} shows examples of feminine terms, but the top masculine terms are often labels of ideological groups, such as \textit{mgtow} and \textit{incels}, which we use to estimate which clusters align with ideological up and downturns.\footnote{MRA gained footing during the height of PUAs, but the peak of \textit{mras}'s frequency is close to the time span's middle.} Cluster A and cluster D tend to have terms that have widespread use. 

\begin{table}[t]
\centering
\resizebox{0.6\columnwidth}{!}{
\begin{tabular}{cc}
\toprule
\textbf{Axis} & \textbf{Variance} \\
\midrule
womanly $\leftrightarrow$ unwomanly & 0.0207  \\
androgynous $\leftrightarrow$ male, female & 0.0105  \\
lovable $\leftrightarrow$ detestable & 0.0104  \\
reputable $\leftrightarrow$ disreputable &  0.0085\\
wholesome $\leftrightarrow$ sickening & 0.0084 \\
clean $\leftrightarrow$ dirty & 0.0078 \\
\bottomrule
\end{tabular}}
\caption{The axes with the largest variance among feminine-leaning terms in \textsc{extreme\_rel}. An extended version of this table with more high-variance axes and examples of top words at each pole is in Appendix~\ref{sec:variance}.}
\label{tbl:mano_var}
\end{table}

\begin{figure}[t]
    \includegraphics[width=\columnwidth]{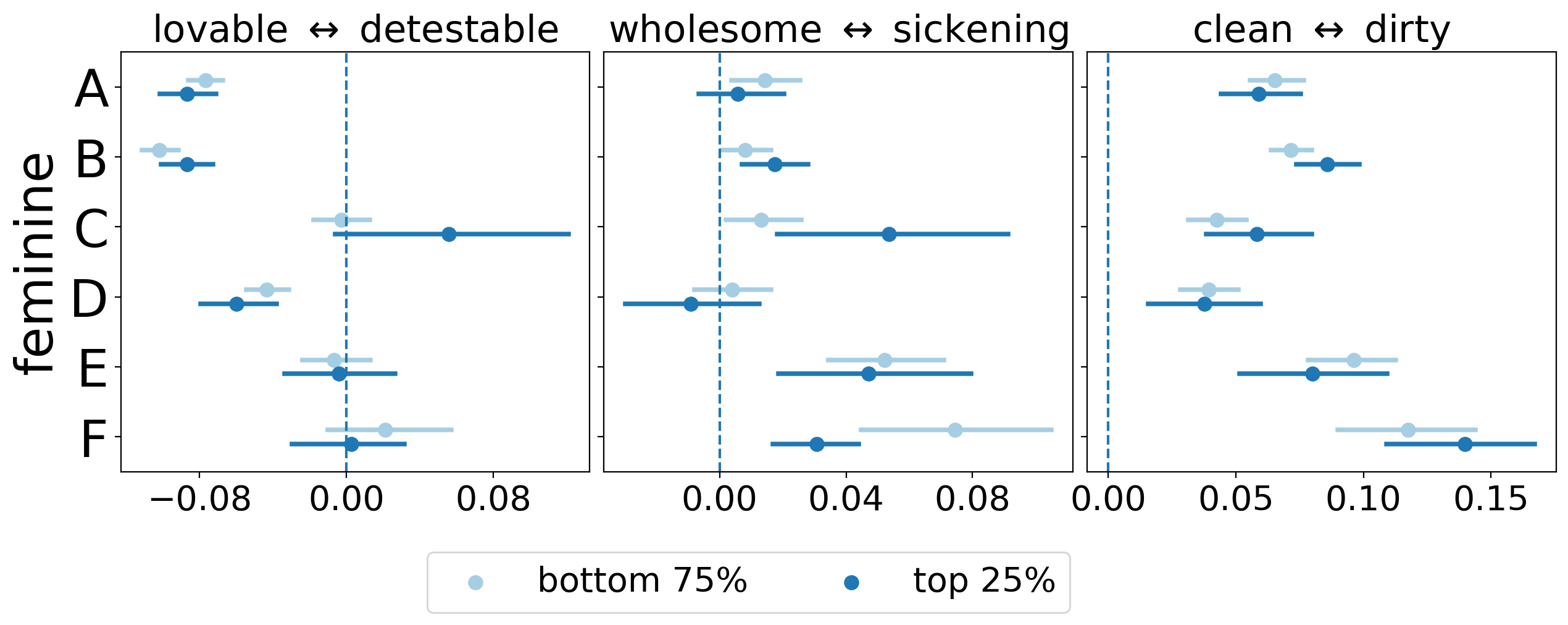}
    \centering
    \caption{Average axis scores among temporal clusters of feminine word types introduced in Figure~\ref{tbl:vocab}. Cluster averages include 95\% CI, vertical dotted lines mark axis midpoints, and clusters are split based on overall frequency percentile. Cluster C, E, and F align with later, more hateful ideologies.}
	\label{fig:cluster_shifts}
\end{figure}

We examine the shifts of high variance, substantive axes across temporal clusters. High variance axes include those related to gender, appearance, and desirability (Table~\ref{tbl:mano_var}). For example, the  \textit{lovable} versus \textit{detestable} pole contrasts \textit{beautiful girls} with \textit{degenerate whores}. As another example, the axis for \textit{clean} versus \textit{dirty} contrasts \textit{loyal wife} with \textit{harlots}. Prior studies using toxicity detection and lexicon-based approaches found that hate and misogyny rose with the arrival of later MGTOW and incel communities \cite{farrell2019exploring,ribeiro2021evolution}. Similarly, we find that lexical choices for women are more \textit{detestable} and \textit{dirty} in later waves associated with MGTOW and incels (Figure~\ref{fig:cluster_shifts}). Often, low and high frequency words share similar patterns in each wave. 

\subsection{Context-level change}\label{sec:mano_context}

\begin{figure*}[t]
    \includegraphics[width=\linewidth]{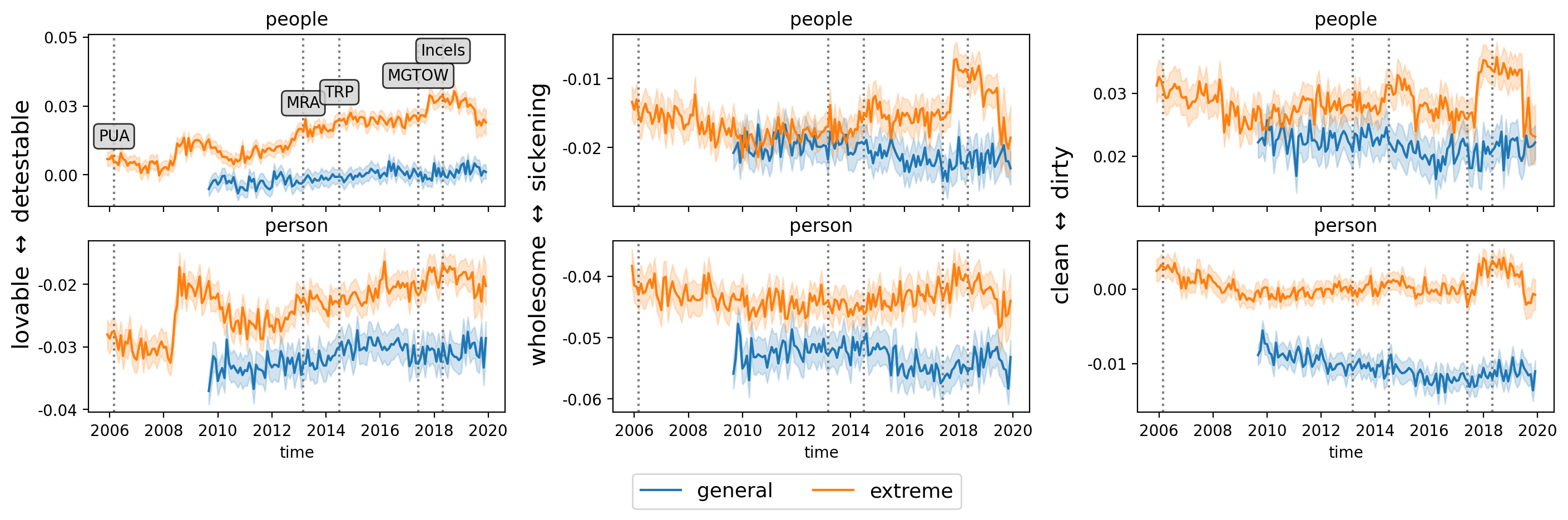}
    \centering
    \caption{Contexts around singular (\textit{person}) or plural (\textit{people}) feminine words over time in \textsc{extreme\_rel} and \textsc{general\_rel} along three axes. Time series include 95\% CI, and dotted lines mark the peak of major ideological communities (gray labels). These vertical lines are months that have the highest normalized frequencies of words used to refer to their members: \textit{puas}, \textit{mras}, \textit{trpers}, \textit{mgtows}, and \textit{incels}.}
	\label{fig:women_over_time}
\end{figure*}

\textbf{Contextualized semantic axes can reveal how the contexts around people have changed over time.} Women in online communities can be referenced in a variety of ways (Figure~\ref{tbl:vocab}). To compare overall changes around women between mainstream and extremist communities, we examine the contexts around feminine (gender-leaning > 0.75) words.  We use instances of 287 unigram types, since bigrams can include modifiers that would be considered ``context''. As discussed earlier, word identities impact measurements of contextual changes across them (Section~\ref{sec:occ_person}). We replace each target word with \textit{person} or \textit{people} depending on whether it is singular or plural, estimated through the Python \textsc{inflect} package. We choose replacements to respect singular/plural forms to ensure ecological validity and not perturb BERT's sensitivity to grammaticality \cite{yin-etal-2020-robustness}. We use reservoir sampling to obtain up to 1000 occurrences of \textit{person}- or \textit{people}-replaced feminine words in each month on \textsc{extreme\_rel} and \textsc{general\_rel}. 

In comparison to \textsc{general\_rel}, \textsc{extreme\_rel} has more \textit{detestable}, \textit{sickening}, and \textit{dirty} contexts for women (Figure~\ref{fig:women_over_time}). Both \textsc{general\_rel} and \textsc{extreme\_rel} discuss relationship issues, but contextualized axes reveal how contrasting and changing attitudes toward women can influence context. Negative associations especially peak during the height of the incels' movement around late 2017 to mid 2019. These persist despite Reddit's ban of r/incels in November 2017 and the quarantine of r/braincels and r/theredpill in September 2018. Thus, the widespread efficacy of community-level moderation is worthy of closer study \citep[e.g.][]{copland2020reddit,ribeiro2021platform}. An advantage of computing scores at the token-level rather than at the type-level is interpretability. That is, one can see which contexts land at the extreme ends of axes (as illustrated in Table~\ref{tbl:context_exp}). 

\begin{table}[t]
\centering
\resizebox{\linewidth}{!}{
\begin{tabular}{p{7.5cm}c}
\toprule
\textbf{Example} & \textbf{Score} \\
\midrule
... use this against us men ... those evil \textbf{people}! & 0.244\\
... these \textbf{people} pollute our public ... & 0.240\\
... parasite worthless whore \textbf{people}. & 0.234\\
\midrule
... I have two little \textbf{people} and they are absolutely amazing ... & -0.156\\
\textbf{people} who are this young and attractive ... & -0.144\\
... my ideal relationship and \textbf{people} like this ... & -0.137\\
\bottomrule
\end{tabular}}
\caption{Examples of \textit{people}, when replacing words for women, in different contexts along the \textit{lovable} $\leftrightarrow$ \textit{detestable} axis in \textsc{extreme\_rel}. These examples have the maximum or minimum score in their month, and were included in the sample used in Figure~\ref{fig:women_over_time}.}
\label{tbl:context_exp}
\end{table}

\textbf{Contextualized semantic axes can also illuminate differences among lexical variables}, or different linguistic forms that share the same referential meaning \cite{nguyen-etal-2021-learning,labov1972sociolinguistic}. As prominent examples, men-led communities use the lexical innovations \textit{femoids} and \textit{foids}, which are shortenings of \textit{female humanoids}, as dehumanizing words for all women \cite{chang2020femoids,prazmo2020foids}. Two women-led communities, Femcels and FDS, use \textit{moids} as an analogous way to refer to men. Prior work studying three manosphere subreddits showed that the lemmas \textit{woman} and \textit{girl} are constructed negatively as immoral, deceptive, incapable and insignificant \cite{krendel2020men}. We hypothesize that the contexts of community-specific variants should have even more dehumanizing connotations along similar dimensions. In this experiment, we replace all terms (\textit{men}, \textit{moids}, \textit{foids}, \textit{femoids}, and \textit{women}) with \textit{people}. 

\begin{figure}[t]
    \includegraphics[width=\columnwidth]{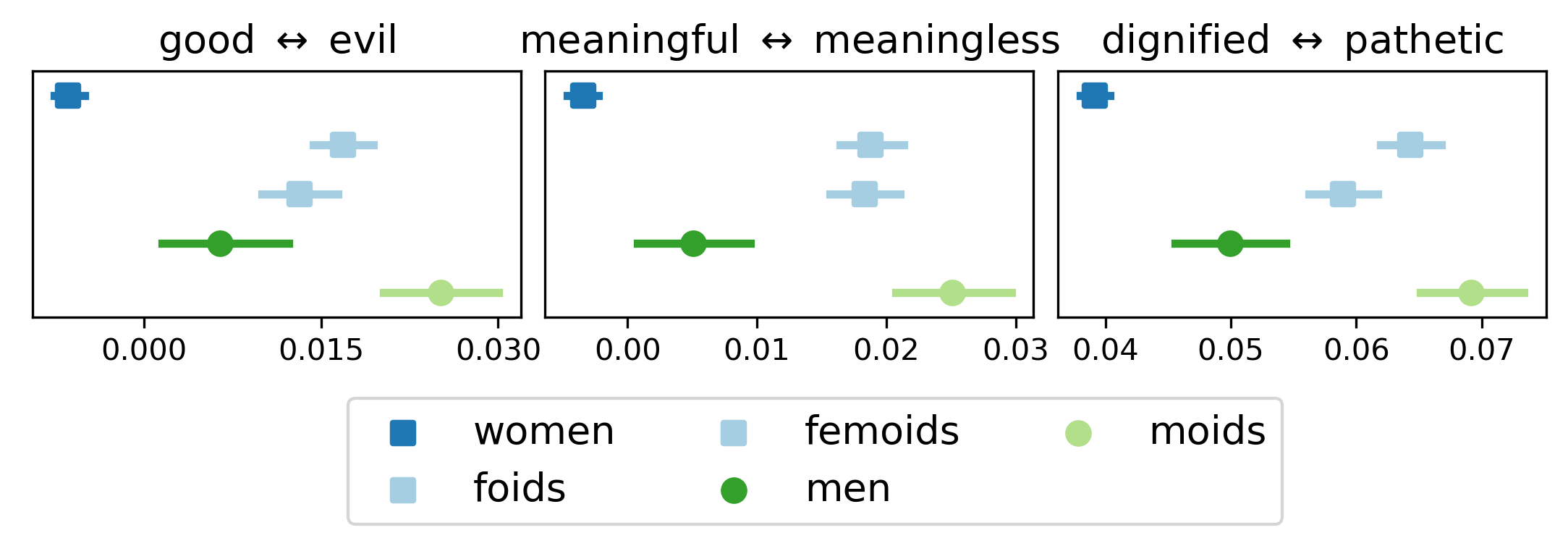}
    \centering
    \caption{Average axis scores, of words used by men-led communities to refer to women (squares), and words used by women-led communities to refer to men (circles). Community-specific variants have lighter colors, and bars indicate 95\% CI.}
	\label{fig:variants}
\end{figure}

We sample up to 100 occurrences of each variant in each platform and ideology per year, limiting time ranges to when domain-specific variants are widely used by their home community. We examine the use of variants for men by Femcels and FDS in 2018-2019, and the use of variants for women by all other communities in \textsc{extreme\_rel} in 2017-2019. Unlike in the \textit{person} experiment for occupations, we have substantial pools of occurrences to compare. Thus, to find axes that distinguish one variant from another, we use axis scores as features in random forest classifiers \cite{scikit-learn}, and perform binary classification of word identity: \textit{women} versus \textit{foids} or \textit{femoids}, and \textit{men} versus \textit{moids} (Appendix~\ref{sec:importance}). We rank axes based on their feature importance, and select three highly ranked and relevant axes to show in Figure~\ref{fig:variants}. Shifts along these axes confirm our hypothesis that community-specific variants are more dehumanized than their widely-used counterparts. 

\section{Conclusion}

In this work, we examine the capability of contextualized embeddings for discovering differences among words and contexts. Our method uses predicted word probabilities to pinpoint which contexts to include when aggregating BERT embeddings to construct axes. This approach creates more self-consistent axes that better fit different occupation categories, in comparison to baselines. We further demonstrate the use of these axes in a longitudinal, cross-platform case study. Overall, contextualized embeddings offer more flexibility and granularity compared to static ones for the analysis of content across time and communities. That is, rather than train static word embeddings for various subsets of data, we can characterize change and variation at the token-level. 

Though we focus on analyzing associations between adjectives and people, our approach can generalize to other types of entities as well. Measuring and comparing the contexts of other entity types should include many of the same considerations we did, such as reducing the conflation of antonyms, controlling for word identity by replacing target words with a shared hypernym, and experimenting with $z$-scoring. Future work includes understanding why some opposing concepts are conflated in large language models, and how a word embedding's identity influences its encoding of contexts.

\section{Limitations}

Aside from computing power requirements (Appendix~\ref{sec:compute}), we outline a few additional limitations of our methodology and its application not discussed in the main text. 

\paragraph{Domain shift.} The use of pretrained BERT on a niche set of communities makes our approaches susceptible to domain shift, such as rare words having less robust embeddings \cite{zhou-etal-2022-richer,zhou2021frequency}, or target words carrying over learned associations from a broader corpus that are less applicable in a narrower one. Domain shift is difficult to avoid without retraining or further pretraining BERT, which is resource-intensive, may risk catastrophic forgetting, and inaccessible to some disciplines in computational social science \cite{gururangan-etal-2020-dont,ramponi-plank-2020-neural,Goodfellow14anempirical}. Also, training a large language model on text with toxic and misogynistic origins introduces additional risk of dual use \cite{kurenkov2022gpt4chan}. We suggest some potential workarounds that lessen the severity of domain shift, such as replacing target words with common ones for context-focused analyses. 


\paragraph{WordNet.} WordNet is a popular lexical resource for NLP, but its senses for words can be overly fine-grained \cite{pilehvar-camacho-collados-2019-wic} and not suitable for all domains. We use WordNet version 3.0, which is included in NLTK, and this version was last updated in 2006. Since English is constantly changing, some synonym and antonym relations may be outdated. 

\paragraph{Errors.} Our method for drawing out differences in words is better than common baselines yet still imperfect, and some of the opposing concepts in embedding space that BERT struggles to separate may be important for an application domain. Therefore, domain expertise is needed to recognize spurious patterns from real ones and fill these gaps. 

In the main text we mention that embeddings offer a ``foggy window'' into how two concepts may be associated or related, and the exact type of relation is not always clear. For example, if contexts for \textit{women} are closer to \textit{unpleasant}, does it mean that the text discusses unpleasant events that affect women, or that the writers believe that women are unpleasant, or both? Some of this uncertainty could be resolved qualitatively by inspecting sentences at poles' extremes. We compare embeddings for people to axes, but it is also possible to include relation-based approaches such as dependency parsing and compare words that share specific relations with people to axes \citep[e.g.][]{lucy-bamman-2021-gender}. One trade-off of doing this is that informative verbs and adjectives connected to mentions of target groups can be sparse. Our method is able to find that \textit{mathematician} replaced with \textit{person} is highly similar to \textit{calculable} in a variety of sentence structures, such as this one modified off Wikipedia: \textit{A \underline{person} is someone who uses an extensive knowledge of mathematics in their work, typically to solve mathematical problems}.

\section{Ethical considerations}

\paragraph{User privacy.} Online data opens many doors for research, but its use raises concerns around user privacy. For our use case, we believe that the benefits of our work outweigh privacy-related harms. Consent is infeasible to obtain for large datasets \cite{buchanan2017ethics}, and in the manosphere, it is unlikely that users would give consent, especially if the researchers using their data believe that their ideologies are harmful and wrong. Obtaining consent would pose risks to the safety of the researcher \cite{conway2021online,doerfler2021dangerous}.

All online discussions included in our work were public when downloaded by their original curators, mainly \citet{baumgartner2020pushshift} and \citet{ribeiro2021evolution}. Some forums and online glossaries were relocated, shutdown, banned, or made private later on. A user's ``right to be forgotten'' confronts researchers who have interests in documenting and studying the histories of communities. We truncate the examples shown in our paper rather than use them in full verbatim \cite{bruckman2002artist}. 

Communities may expect their posts to stay within their in-group, but the content in our work was posted on public platforms. This publicness and increased visibility plays a key role in how this content impacts others, such as those who view this information and propagate it elsewhere, or those who are direct targets of hate. Common targets such as women and people of color carry a bigger burden when participating in online spaces \cite{hoffman2017recasting}, and our broader research agenda aims to mitigate this issue. 

\paragraph{Social biases in models and resources.} We use WordNet to group similar adjectives into semantic axes, but we observe some socially harmful associations in this resource. For example, \textit{gross} and \textit{fat} are listed as similar lemmas. As another example, WordNet conflates gender and sexuality when \textit{androgynous} and \textit{bisexual} are also listed as similar lemmas. The BERT language model, like all large, pretrained models, is also susceptible to social biases in its training data \cite{bender2021dangers}.

\paragraph{Gender inference.} In this paper's main case study, we perform gender inference for word and phrase types. This step was necessary to study how women are portrayed over time, which is a key question due to the centrality of misogyny in these communities. However, perfect prediction of each word's perceived gender in our dataset using pronouns is impossible \cite{cao-daume-iii-2021-toward}. Not all mentions of people co-occur with pronouns, pronouns do not equate gender, and coreference resolution systems can produce errors. So, we approximate the social gender of terms by aggregating coreference patterns over all instances of that term. Since it is difficult to separate noisy errors from meaningful word-level pronoun variation at scale, we had to use a score threshold to pinpoint what words were feminine-leaning enough to be included in our analyses. 

Restricting pronouns to the traditional binary of feminine and masculine is limiting, since individuals use other pronouns as well. \textit{They}/\textit{them} pronouns are predominantly used to reference plural terms in this dataset, and the coreference model we use does not handle neopronouns. The manosphere and the typical framing under which it is studied is heavily cisheteronormative. We use a frequency cutoff to determine our vocabulary (Appendix~\ref{sec:appdx_vocab}), so references to transgender and nonbinary people may be filtered out. Vocab terms retained for transgender people are outdated or typically offensive terms such as \textit{transsexuals} and \textit{transgenders}, and no vocab term includes \textit{non-binary}, \textit{nb}, or \textit{nonbinary}. 

\section{Acknowledgements}

We thank Manoel Horta Ribeiro for sharing his dataset and materials for our case study, and Sam Robertson, Alexus Lopez, and Harold Cha for evaluating model outputs. In addition, we are grateful for feedback provided by Nicholas Tomlin, Kaitlyn Zhou, and our anonymous reviewers. This research was supported by funding from the National Science Foundation (DGE-1752814, IIS-1813470, and IIS-1942591). 

\bibliography{anthology,custom}
\bibliographystyle{acl_natbib}

\appendix

\section{Wikipedia page titles}\label{appdx:wiki} 

Table~\ref{tbl:occ_cats} lists the categories of occupations, the titles of Wikipedia pages that list them, and the number of terms in each category. These lists were retrieved in February 2022. 

\section{Human evaluation for occupations}
\label{sec:human_eval}

\begin{figure}[t]
    \fbox{\includegraphics[width=0.9\linewidth]{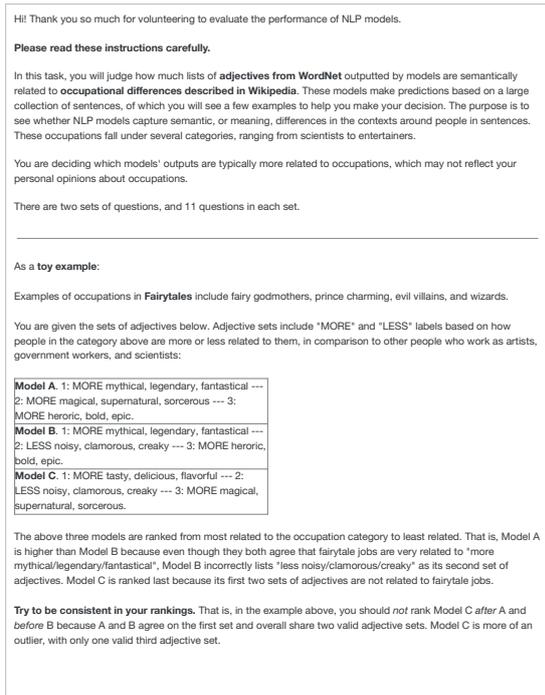}}
    \centering
    \caption{Instructions and a toy example shown to human evaluators.}
	\label{fig:human_eval_instruct}
\end{figure}

We recruited three student volunteers with familiarity with NLP coursework and tasks to rank the top poles provided by each axis-building method for our occupation and \textit{person} experiments. We used Qualtrics to design and launch the survey. Since we were not asking about personal opinions but rather evaluating models, we were determined exempt from IRB review by the appropriate office at our institution. Each question pertains to a specific occupation category, and within each experiment, question order and answer option order are randomly shuffled. Each model option is presented with its top three poles, in order of most to less relevant. Figure~\ref{fig:human_eval_instruct} shows screenshots of instructions. In the toy example, the options are labeled with ``Model A", ``Model B", ``Model C", to allow explanation clarity, but in the actual task questions, options are not labeled with model letters to avoid biasing the evaluators towards a specific model. Some annotators expressed that the task was difficult, and for some occupations, different approaches output similar axes, just in different order.

\begin{table}[t]
\centering
\resizebox{\columnwidth}{!}{
\begin{tabular}{p{4cm}p{7cm}c}
\toprule
\textbf{Occupation type} & \textbf{Wikipedia category lists} & \textbf{\# of terms} \\
\midrule
Writing & ``List of writing occupations" & 27 \\
Entertainment & ``List of theatre personnel", ``List of film and television occupations" & 48 \\
Art &  ``List of artistic occupations" & 32 \\
Health &  ``List of healthcare occupations" & 47 \\
Agriculture & ``Category:Agricultural occupations", and plant and husbandry subcategory & 29 \\
Government & ``Category:Government occupations" & 47 \\
Sports &  ``Lists of sportspeople"& 24 \\
Engineering  &  ``List of scientific occupations"& 7 \\
Physical, natural, and earth sciences (``Science'') &  ``List of scientific occupations"& 30 \\
Math \& statistics &  ``List of scientific occupations"& 3 \\
Social sciences &  ``List of scientific occupations"& 6 \\
\bottomrule
\end{tabular}}
\caption{Wikipedia page titles for pages containing lists of occupations.}
\label{tbl:occ_cats}
\end{table}

\section{Reddit communities}
\label{sec:appdx_communities}

We used a list of subreddits\footnote{This list can be found at our Github repo: \href{https://github.com/lucy3/context_semantic_axes/blob/master/data/subreddits.txt}{https://github.com/lucy3/context\_semantic\_axes}.} for the manosphere provided by \cite{ribeiro2021evolution} in their detailed, data-driven sketch of the manosphere. 

Five of the subreddits included in \citet{ribeiro2021evolution}'s taxonomy of the Reddit manosphere (r/malecels, r/lonelynonviolentmen, r/1ncels, r/incelbrotherhood, r/incelspurgatory) were not on Pushshift's dump of Reddit. We curated the list of communities for our new ideological category, Female Dating Strategy (FDS), using a now removed list of FDS's ``sister communities" on the subreddit r/FemaleDatingStrategy's sidebar: r/PinkpillFeminism, r/AskFDS, r/FDSSuperFans, r/PornFreeRelationships, and r/FemaleLevelUpStrategy. The Femcels set of subreddits include: r/Trufemcels, r/TheGlowUp, and r/AskTruFemcels. Though the main user base of the manosphere are men, there are also small populations of women in other ideologies as well, such as r/redpillwomen. We mainly portion out FDS and Femcels due to their role in Section~\ref{sec:mano_context}'s lexical variant experiment as communities who use \textit{moids}. 

In total we have 12 subreddits in TRP, 11 in MRA, 7 in PUA, 22 in Incels, 3 in MGTOW, 4 in Femcels, and 6 in FDS. The complete list of subreddits and their categories is also in our Github repo. 

\section{Vocabulary creation}
\label{sec:appdx_vocab}

First, we extract nominal and proper persons using NER, keeping ones that are popular (occur at least 500 times in \textsc{extreme\_rel}), and unambiguous, where at least 20\% of its instances in these datasets are tagged as a person. Gathering a substantial number of labels from our domain to train an in-domain NER system from scratch is outside the scope of our work, so we experimented with three models trained on other labeled datasets: ACE, contemporary literature, and a combination of both. We evaluated these models on a small set of posts and comments labeled by one author, after retrieving 25 examples per forum or Reddit ideological category using reservoir sampling. The annotator only labeled spans for nominal and named \textsc{Person} entities. Table~\ref{tbl:ner} shows the performance of each model on \textsc{extreme\_rel}. Based on these evaluation results, we chose to use the model trained on contemporary literature. 

We extract bigrams and unigrams from detected spans, excluding determiners and possessives whose heads are the root of the span. Named entities that refer to types of people rather than specific individuals were estimated through their co-occurrence with the determiner \textit{a}, e.g. \textit{a Chad}.

\begin{table}[t]
\centering
\resizebox{0.8\columnwidth}{!}{
\begin{tabular}{lccc}
\toprule
\textbf{Training data} & \textbf{Precision} & \textbf{Recall} & \textbf{F1}\\
\midrule
ACE & 0.657 & 0.641 &  0.649\\
Literature & 0.798 & 0.711 & \textbf{0.752}\\
Combined & 0.715 & 0.744 & 0.729\\
\bottomrule
\end{tabular}}
\caption{Model performance on a human-annotated sample of Reddit and forum data. The F1 score we used to determine our choice of model is highlighted in bold.}
\label{tbl:ner}
\end{table}

Then, one author consulted community glossaries and examined in-context use of words to manually correct the list of automatically extracted terms. We include additional popular and unambiguous words not tagged sufficiently often enough by NER, but defined as people in prior work and online resources. 

Table~\ref{tbl:gloss} lists the sources and glossaries for vocabulary words and the ideologies they include. Some of these sources, such as the Shedding of the Ego, are created by insiders in the community, while some, such as academic papers and news articles, are by outsiders. For each of these glossaries and lists of terms, we manually separated them into two categories: 269 people (singular and plural forms) and 1776 non-people. Two of these sites, Shedding of the Ego and Pualingo, no longer exists, but were publicly available until at least late 2020. We include 93 terms for people that were initially filtered out in our NER pipeline in our final vocabulary, excluding ambiguous ones that also occur often as non-human entities, such as \textit{tool} (a fool who is taken advantage of) and \textit{jaw} (short for \textit{just another wannabe}). 

The resulting vocabulary contains niche language, where 20.7\% of unigrams are not found in WordNet, and 85.1\% of those missing are also not in the Internet resource Urban Dictionary.\footnote{We use the naive approach of adding or removing `-s' to search for either plural or singular forms in these lexicons.} The full list is also available in our Github repo. 

\begin{table}[t]
\centering
\resizebox{\columnwidth}{!}{
\begin{tabular}{lcc}
\toprule
\textbf{Source} & \textbf{Medium} & \textbf{Community} \\
\midrule
\citet{jaki2019incels} & academic paper & Incels\\
\citet{ging2019masculinities} & academic paper & Manosphere \\
\citet{farrell2019exploring} & academic paper & Manosphere \\
\citet{Lin2017antifem} & academic paper & MGTOW\\
\citet{squirrell2018} & blog & Incels\\
\citet{sonnad2017}& news & Red Pill\\
\citet{beran2018} & magazine & Incels \\
Pualingo & website & PUA\\
Rational Wiki & website & Manosphere\\
Shedding of the Ego & blog & MGTOW, PUA\\
\bottomrule
\end{tabular}}
\caption{Sources for non-NER detected terminology we include in our study. Shedding of the Ego can be viewed in the \href{https://web.archive.org/web/20190827053903/http://sheddingoftheego.com/}{Internet Archive}. On the other hand, Pualingo was taken down and removed from the Internet archive during the preparation of this paper. In some cases, the focus community is the entire manosphere, while in others, it is a subset.}
\label{tbl:gloss}
\end{table}

\section{Gender inference} \label{sec:gender_appdx}

This section includes additional details around our gender inference process. 

Our list of semantically gendered terms, or words gendered by definition, expands upon the one used by \citet{hoyle-etal-2019-unsupervised}: \textit{man, men, boy, boys, father, fathers, son, sons, brother, bothers, husband, husbands, uncle, uncles, nephew, nephews, emperor, emperors, king, kings, prince, princes, duke, dukes, lord, lords, knight, knights, waiter, waiters, actor, actors, god, gods, policeman, policemen, postman, postmen, hero, heros, wizard, wizards, steward, stewards, woman, women, girl, girls, mother, mothers, daughter, daughters, sister, sisters, wife, wives, aunt, aunts, niece, nieces, empress, empresses, queen, queens, princess, princesses, duchess, duchesses, lady, ladies, dame, dames, waitress, waitresses, actress, actresses, goddess, goddesses, policewoman, policewomen, postwoman, postwomen, heroine, heroines, witch, witches, stewardess, stewardesses}. 

We include the following additional semantically gendered terms: \textit{male, males, dude, dudes, guy, guys, boyfriend, boyfriends, bf, female, females, chick, chicks, girlfriend, girlfriends, gf, gal, gals, bro, transmen, transwomen, she, he}. 

We check if any of the above words appear in a unigram or bigram vocabularly term. Around 29.9\% of our vocabulary in \textsc{extreme\_rel} is gendered through this word list approach. 

\begin{table*}[t]
\centering
\resizebox{\linewidth}{!}{
\begin{tabular}{ccp{15cm}}
\toprule
\multicolumn{3}{c}{\textbf{Feminine}} \\
\midrule
\textbf{Axis} & \textbf{Variance} & \textbf{Examples} \\
\midrule
womanly $\leftrightarrow$ unwomanly & 0.0207 & \textit{female gender, feminine women, feminine woman} $\leftrightarrow$ \textit{hambeast, tomboys, tomboy} \\
androgynous $\leftrightarrow$ male, female & 0.0105 & \textit{manipulative bitch, nympho, noodlewhore} $\leftrightarrow$ \textit{white females, female, females} \\
lovable $\leftrightarrow$ detestable & 0.0101 & \textit{little princess, sweet girl, beautiful girl} $\leftrightarrow$ \textit{stupid cunts, degenerate whores, accusers} \\
reputable $\leftrightarrow$ disreputable & 0.0085 & \textit{great wife, great woman, great women} $\leftrightarrow$ \textit{slut, dirty slut, sluts}\\
wholesome $\leftrightarrow$ sickening & 0.0084 & \textit{homemakers, healthy woman, healthy women} $\leftrightarrow$ \textit{evil bitch, dirty slut, degenerate whores}\\
clean $\leftrightarrow$ dirty & 0.0078 & \textit{loyal wife, healthy woman, perfect woman} $\leftrightarrow$ \textit{club sluts, hambeasts, harlots}\\
conventional $\leftrightarrow$ unconventional & 0.0076 & \textit{most women, average female, female counterparts} $\leftrightarrow$ \textit{debbie downer, sissy, fuckbuddy} \\
beautiful $\leftrightarrow$ ugly & 0.0075 & \textit{great girl, beautiful wife, gorgeous girl} $\leftrightarrow$ \textit{fat pig, female rapist, degenerate whores} \\
proud $\leftrightarrow$ humble & 0.0070 & \textit{harlots, manipulative bitch, harlot} $\leftrightarrow$ \textit{zero women, few women, most females}\\
competent $\leftrightarrow$ incompetent& 0.0069 & \textit{female lawyer, good pussy, female judge} $\leftrightarrow$ \textit{harlots, degenerate whores, unattractive woman}\\
old $\leftrightarrow$ young & 0.0069 & \textit{old hags, old hag, old ex} $\leftrightarrow$ \textit{young teen, toddler, toddlers}\\
\bottomrule
\end{tabular}}
\caption{An extended version of Table~\ref{tbl:mano_var}. The axes with the largest variance among feminine-leaning terms in \textsc{extreme\_rel}. Examples shown are the top three for each pole.}
\label{tbl:mano_var_extended}
\end{table*}

To infer gender for the remaining words using pronouns, we ran coreference resolution on \textsc{extreme\_rel}, and extracted all pronouns that are clustered in coreference chains with terms in our vocabulary \cite{clark-manning-2016-deep}. We label the masculine to feminine leaning of vocab terms by calculating the proportion of feminine pronouns (\textit{she}, \textit{her}, \textit{hers}, \textit{herself}) over the sum of feminine and masculine pronouns (\textit{he}, \textit{him}, \textit{his}, \textit{himself}). We only consider a word to have a usable gender signal if it appears in at least 10 coreference clusters with feminine or masculine pronouns. Since plural words do not usually appear with \textit{he}/\textit{she} pronouns, we have plural words take on the gender leaning of their singular forms. We pair plural and singular forms using the Python \textsc{inflect} package.\footnote{https://pypi.org/project/inflect/} We also transfer unigrams' gender to bigrams, after examining the modifiers (the first token) in bigram terms to check that they are not differently and semantically gendered.  Around 20.9\% of our vocabulary in \textsc{extreme\_rel} is gendered through pronouns alone, an additional 12.6\% is gendered through plural to singular mapping, and an additional 9.1\% is gendered through bigram to unigram mapping.

\section{High variance axes} \label{sec:variance}

Table~\ref{tbl:mano_var_extended} shows the top vocabulary terms that correspond to the poles of high variance axes. 

\section{Classification of lexical variants} \label{sec:importance}

Our main goal here is to tease out which axes differentiate the contexts of lexical variants, rather than find the best model that performs well on a classification task. Therefore, we choose to use a random forest classifier for its interpretability: it outputs weights that indicate what features were most important across its decisions. We use scikit-learn's implementation, and perform randomized search with 5-fold cross validation and weighted F1 scoring to select model parameters (Table~\ref{tbl:params}). Table~\ref{tbl:forest_importance} shows the most important axis features of these models. In general, the set of most important features did not change much with parameter choices and roughly aligns with axes that showcase the largest mean differences between each pair of variants. That is, the three axes we show in the main text in Figure~\ref{fig:variants} are also among the top ten ordered by mean difference for \textit{men} vs. \textit{moids} and \textit{women} vs. \textit{femoids}. 

\section{Runtime and infrastructure} \label{sec:compute}

We only use BERT-base for inference, but the overall runtime cost is high due to the size of our corpora: English Wikipedia and social media discussions. We use one Titan XP GPU with 8 CPU cores for most of the paper, and occasionally expanded to multiple machines with 1080ti and K80 GPUs in parallel when handling social media data. We use BERT for two main purposes: predicting word probabilities to select contexts for constructing axes, and obtaining word embeddings. On one Titan XP GPU, the former takes $\sim$1 hour for one million sentences containing one masked target word each, and the latter takes $\sim$2.5 hours for one million sentences, including wordpiece aggregation. 

\begin{table}[t]
\centering
\resizebox{0.6\columnwidth}{!}{
\begin{tabular}{ll}
\toprule
\textbf{Parameter} & \textbf{Choices} \\
\midrule
n\_estimators & 50$^*$, 100, 150$^{\dagger\ddag}$, 200 \\ 
criterion& entropy$^{\dagger\ddag}$, gini$^*$\\ 
max\_depth & None, 10$^*$, 50, 70$^\dagger$, 100$^\ddag$\\ 
max\_features & auto$^\ddag$, sqrt$^{\dagger*}$\\ 
min\_samples\_split & 2$^\dagger$, 5$^{\ddag*}$, 10, 20\\ 
min\_samples\_leaf & 1$^\ddag$, 2$^{\dagger*}$, 4\\ 
\bottomrule
\end{tabular}}
\caption{Parameter choices for random forest classification. Symbols mark selected parameters for each task, where $\dagger$ refers to \textit{men} vs. \textit{moids}, $\ddag$ refers to \textit{women} vs. \textit{femoids}, and $*$ refer to \textit{women} vs. \textit{foids}. These models had weighted F1 scores of 0.670, 0.759, and 0.781, respectively.}
\label{tbl:params}
\end{table}

\begin{table}[t]
\centering
\resizebox{0.6\columnwidth}{!}{
\begin{tabular}{cc}
\toprule
\textbf{Axis} & \textbf{Importance} \\
\midrule
\multicolumn{2}{c}{\textit{men} vs. \textit{moids}} 
\\\cmidrule(lr){1-2}
violent $\leftrightarrow$ nonviolent &  0.0078\\ 
useful $\leftrightarrow$ useless & 0.0064\\ 
possible $\leftrightarrow$ impossible & 0.0064\\ 
wholesome $\leftrightarrow$ sickening & 0.0063\\ 
\textbf{meaningful $\leftrightarrow$ meaningless} & 0.0055\\ 
\midrule
\multicolumn{2}{c}{\textit{women} vs. \textit{femoids}} 
\\\cmidrule(lr){1-2}
lost (e.g. \textit{doomed}) $\leftrightarrow$ saved & 0.0081 \\ 
\textbf{dignified $\leftrightarrow$ pathetic} & 0.0073 \\ 
\textbf{good $\leftrightarrow$ evil} & 0.0071 \\ 
\textbf{meaningful $\leftrightarrow$ meaningless} & 0.0070 \\ 
high $\leftrightarrow$ low & 0.0063 \\ 
\midrule
\multicolumn{2}{c}{\textit{women} vs. \textit{foids}} 
\\\cmidrule(lr){1-2}
empirical $\leftrightarrow$ theoretical & 0.0112 \\ 
\textbf{good $\leftrightarrow$ evil} & 0.0101 \\ 
blond $\leftrightarrow $ brunet& 0.0095 \\ 
\textbf{meaningful $\leftrightarrow$ meaningless}& 0.0090\\ 
shapely $\leftrightarrow$ unshapely& 0.0084 \\ 
\bottomrule
\end{tabular}}
\caption{Feature importances in random forest classifiers that predict the identity of a target word, where features are words' axes scores. Bolded axes are featured in the main text in Figure~\ref{fig:variants}.}
\label{tbl:forest_importance}
\end{table}

\end{document}